\documentclass[sigconf]{acmart}

\usepackage{graphicx}
\usepackage{amsmath,amssymb} 
\usepackage{color}
 \usepackage{balance}
\usepackage{float}
\usepackage{booktabs} 
\usepackage{multirow}

\usepackage{subfigure}

\copyrightyear{2018}
\acmYear{2018}
\setcopyright{acmcopyright}

\acmConference[MM '18]{2018 ACM Multimedia Conference}{October 22--26, 2018}{Seoul, Republic of Korea}
\acmBooktitle{2018 ACM Multimedia Conference (MM '18), October 22--26, 2018, Seoul, Republic of Korea}
\acmPrice{15.00}
\acmDOI{10.1145/3240508.3240525}
\acmISBN{978-1-4503-5665-7/18/10}
\begin{document}
\title{Attention-based Pyramid Aggregation Network \\for Visual Place Recognition}

\author{Yingying Zhu}
\affiliation{%
\institution{College of Compute Science and Software Engineering, Shenzhen University, China}
}
\email{zhuyy@szu.edu.cn}

\author{Jiong Wang}
\affiliation{%
  \institution{College of Compute Science and Software Engineering, Shenzhen University, China}
}
\email{liubinggunzu@gmail.com}

\author{Lingxi Xie}
\affiliation{\institution{Department of Computer Science, Johns Hopkins University, USA}
}
\email{198808xc@gmail.com}

\author{Liang Zheng}
\affiliation{\institution{Research School of Computer Science, Australian National University, AUS}
}
\email{liangzheng06@gmail.com}

\begin{abstract}
	
Visual  place recognition is challenging in the urban environment and is usually viewed as a large scale image retrieval task. 
The intrinsic challenges in place recognition exist that the confusing objects such as cars and trees frequently occur in the complex urban scene, and buildings with repetitive structures may cause over-counting and the burstiness problem degrading the image representations.
%
To address these problems, we present an Attention-based Pyramid Aggregation Network (APANet), which is trained in an end-to-end manner for place recognition.  One main component of APANet, the spatial pyramid pooling, can effectively encode the multi-size buildings containing geo-information. 
The other one, the attention block, is adopted as a region evaluator for suppressing the confusing regional features while highlighting the discriminative ones. 
When testing, we further propose a simple yet effective PCA power whitening strategy, which significantly improves the widely used PCA whitening by reasonably limiting the impact of over-counting. 
Experimental evaluations demonstrate that the proposed APANet outperforms the state-of-the-art methods on two place recognition benchmarks, and generalizes well on standard image retrieval datasets. 

\end{abstract}

%
%



\maketitle

	\section{Introduction}

	Visual place recognition has received a considerable level of attention in the community for its wide applications in augmented reality \cite{middelberg2014scalable,chen2009streaming}, autonomous navigation \cite{mcmanus2014shady,hays2008im2gps} and 3D reconstruction \cite{agarwal2011building,crandall2011discrete}. Traditionally, visual place recognition has been cast as an  image retrieval task at city-scale. Given a query image depicting the scene of a particular location, we aim to find most similar images as location suggestions by querying a large geo-tagged database. 
	
\begin{figure}[t]
		\setlength{\abovecaptionskip}{0.cm}
		\setlength{\belowcaptionskip}{-0.cm}
		\centerline{
			\includegraphics[scale = 0.57]{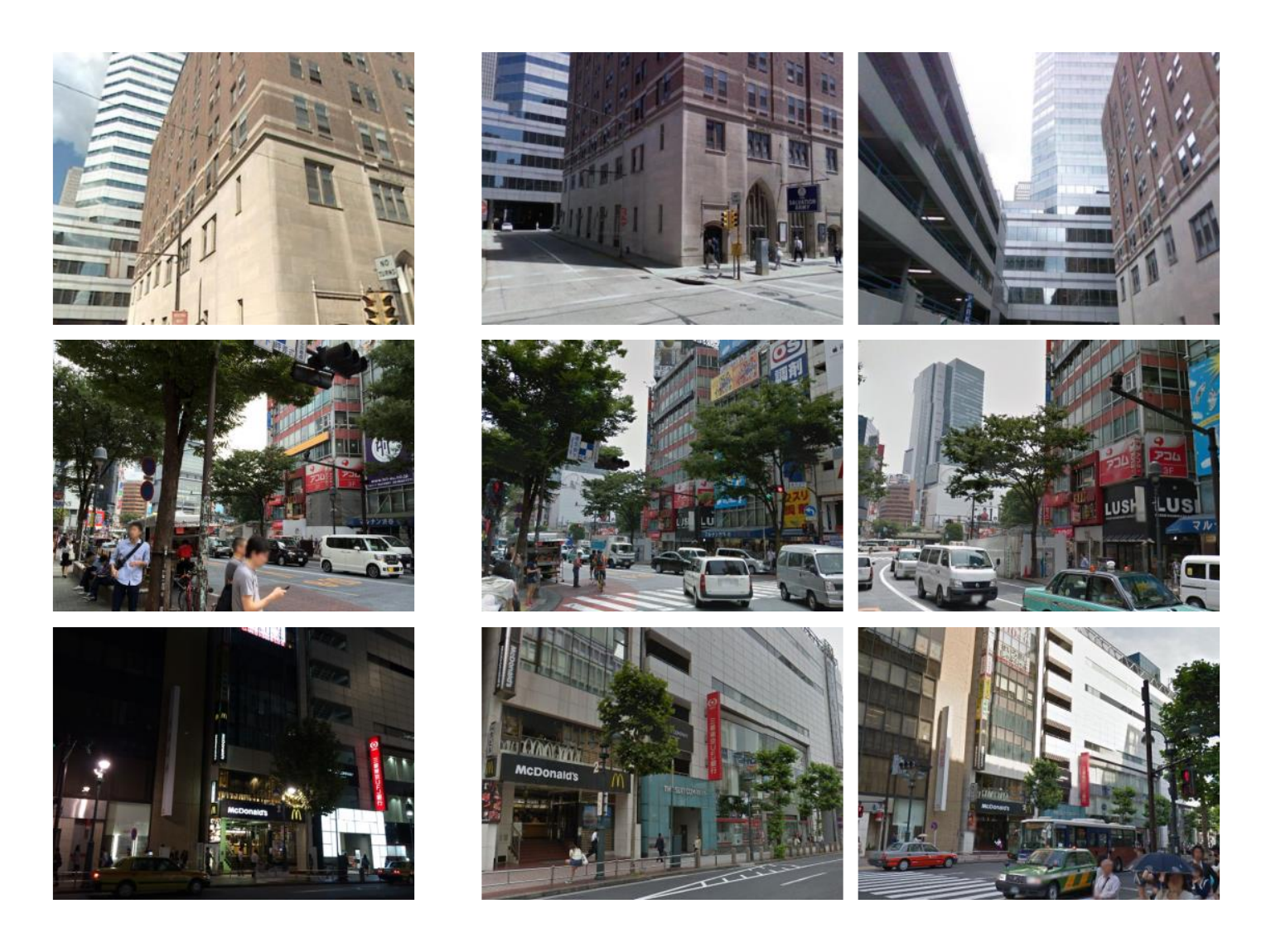}
		}
		\caption{Retrieval examples from place recognition datasets. Left: query images. Right: retrieved database images of the same places. 
		}
		\label{image_example}
	\end{figure}

	Visual place recognition focuses on the complex urban environment, which contains buildings with repetitive structures \cite{torii2013visual} and suffers from changes of illumination conditions, seasons or structural modifications over time \cite{torii201524}. Some representative retrieval examples are shown in Figure \ref{image_example}. An inherent problem exists that the trees and cars frequently occur in the urban environment to cause confusion. 
	The buildings (which are geo-informative) with repetitive structures may also cause the problem of over-counting and burstiness \cite{jegou2009burstiness}, \emph{i.e.,} similar descriptors appear such much times in an image as to degrade the image representation. 
	Moreover, partial occlusions and the changes of viewpoint make place recognition a challenging task.  
	
	Conventional image retrieval techniques such as bag-of-visual-words (BOW) representation \cite{sivic2003video} based on local invariant features 
	\cite{lowe2004distinctive} and vector of locally aggregated descriptors (VLAD) \cite{jegou2012aggregating} had been  adopted to build accurate place recognition systems \cite{knopp2010avoiding,arandjelovic2014dislocation,torii2013visual,torii201524}.  
	An innovative work \cite{arandjelovic2016netvlad} learned discriminative convolutional neural network (CNN) representations on the Street View training datasets and proposed the NetVLAD layer that implements VLAD by series of differentiable operations. The NetVLAD representations yield competitive results on place recognition and image retrieval datasets. 
To suppress the confusing elements,  Kim \emph{et al.} \cite{kim2017learned} extend NetVLAD by learning the contextual weights for the local CNN features. But these contextual weights may act as the burstiness frequencies, thus being restricted by  intra-normalization \cite{Arandjelovic2013All} in the NetVLAD layer. [24] reports similar performances to NetVLAD on Pitts250k-test dataset \cite{torii2013visual} and Tokyo24/7 daytime subset \cite{torii201524}. Therefore [24] has not effectively alleviated the influence of confusing objects in the urban environment. In addition, the NetVLAD layer produces high-dimensional representations (\emph{e.g.,} $64 \times512$ for VGGNet \cite{simonyan2014very}) and requires a large PCA whitening matrix for dimensionality reduction.

Toward accurate and fast place recognition, we propose an attention-based pyramid aggregation network (APANet), an end-to-end trainable model, which consists of three additional blocks on the base CNN architectures: the spatial pyramid pooling block, attention block, and sum pooling block. 
	As shown in Figure~\ref{SPA}, the spatial pyramid pooling is employed to aggregate the multi-size regions on the CNN feature maps. We found this region-based pooling method is effective for encoding the multi-size buildings with repetitive architecture than the global pooling methods \cite{babenko2015aggregating,razavian2016visual}.
	The attention block helps to weight the regional features according to the distinctiveness and then sum pooling produces a compact global descriptor.  
   
    In a nutshell, the attention mechanism can be regarded as a strategy that selectively focuses on the informative visual elements, similar to the human perception process. Recently it was introduced to deep neural networks as a powerful addition in a series of vision tasks  \cite{wang2017residual,hu2018senet,Yan2017Fine}. 
    Inspired by these successful applications, we incorporate the attention block to suppress the confusing elements in the urban environment. Specifically, we employ two kinds of attention blocks, a single attention block and a cascaded attention block with content prior, to weight the regional features by their distinctiveness.  The discriminative regional features will be assigned higher weights than the confusing ones, so that they contribute more to the global descriptor. 
	
	%
	In testing stage, we found the widely used PCA whitening approach address the over-counting in an extreme way. Therefore we develop a PCA power whitening strategy which reasonably address the problem of over-counting to get a maximum level of improvement on the retrieval performance.
	Different with recent works \cite{babenko2014neural,radenovic2016cnn} which learn discriminative dimension reduction and whitening using labeled image pairs, PCA power whitening is in a fully unsupervised way and consistently improves PCA whitening without extra computations.
	
	Our APANet is trained in an end-to-end manner on the Street View training datasets \cite{arandjelovic2016netvlad} targeting for place recognition. 
	On two place recognition benchmark datasets,  APANet representations outperform NetVLAD on the same representation dimensionality especially when the  dimensionality is low. On the standard image retrieval datasets, APANet surpasses NetVLAD with 8-times more compact representations.

	\section{Related Work}
	\subsection{Visual Place Recognition.}
            
	Visual Place Recognition in the urban environment is a challenging task due to the frequently occurred confusing objects, the repetitive structures, changes of viewpoint and illumination condition.	
	Based on the bag-of-features model, some previous works \cite{knopp2010avoiding,arandjelovic2014dislocation} focused on discovering distinctive and confusing local descriptors, thereby exploiting selecting  or weighting strategies. 
	Torii \emph{et al.} \cite{torii2013visual}  explicitly detected the repetitive image structures and developed an efficient representation of the repeated structures for place recognition in the urban environments. In \cite{torii201524}, view synthesis is combined with densely sampled VLAD descriptors to enable robust recognition against  variations in viewpoint and illumination.
	
	
	Arandjelovi{\'c} \emph{et al.} \cite{arandjelovic2016netvlad} performed learning for place recognition and proposed the NetVLAD representations, which significantly outperform the local-feature-based representations on place recognition benchmarks. However,  NetVLAD representations may be degraded by the confusing objects and  need a large PCA whitening matrix for dimensionality reduction. In contrast, the proposed APANet produces compact representations and the built-in attention blocks can effectively suppress the confusing objects. 
	
	
	\subsection{CNN-based Image Retrieval.}
	The seminal work \cite{krizhevsky2012imagenet} has discussed the feasibility of CNN features for image retrieval. \cite{sharif2014cnn,babenko2014neural} extracted CNN activations from the fully-connected (FC) layer as global descriptors and got preliminary results for image retrieval. However, extracting a single feature vector from the FC layer requires a fixed input image size. Subsequent works exploited the global pooling  \cite{babenko2015aggregating,razavian2016visual} or region-based pooling methods \cite{tolias2015particular} on activations of the intermediate convolutional layer. Among these works, 
	Tolias \emph{et al.} \cite{tolias2015particular} propose the R-MAC descriptor that aggregates the regional features generated by three scale rigid grids. The regional features are  $L_2$-normalized, then PCA whitened and  $L_2$-normalized again before sum-aggregation.   
	In combination with re-ranking and query expansion \cite{chum2007total,chum2011total}, R-MAC reports competitive performance. 
	%
	
	The above-mentioned works adopt pre-trained CNNs while recent state-of-the-art works focus on fine-tuning the pre-trained CNNs on the domain-specific datasets \cite{babenko2014neural,arandjelovic2016netvlad,radenovic2016cnn}. A pioneering work \cite{babenko2014neural} fine-tuned CNN models  on a collected Landmark dataset with cross-entropy loss, which improves the retrieval performance a lot.  More recent works \cite{arandjelovic2016netvlad,gordo2016deep,zheng2017sift}  reveal the effectiveness of the triplet ranking loss for CNNs fine-tuning  in image retrieval and place recognition task.
	
	
		\begin{figure*}[ht]
		\setlength{\abovecaptionskip}{0.2cm}
		\centerline{
			\includegraphics[width=1.0\linewidth]{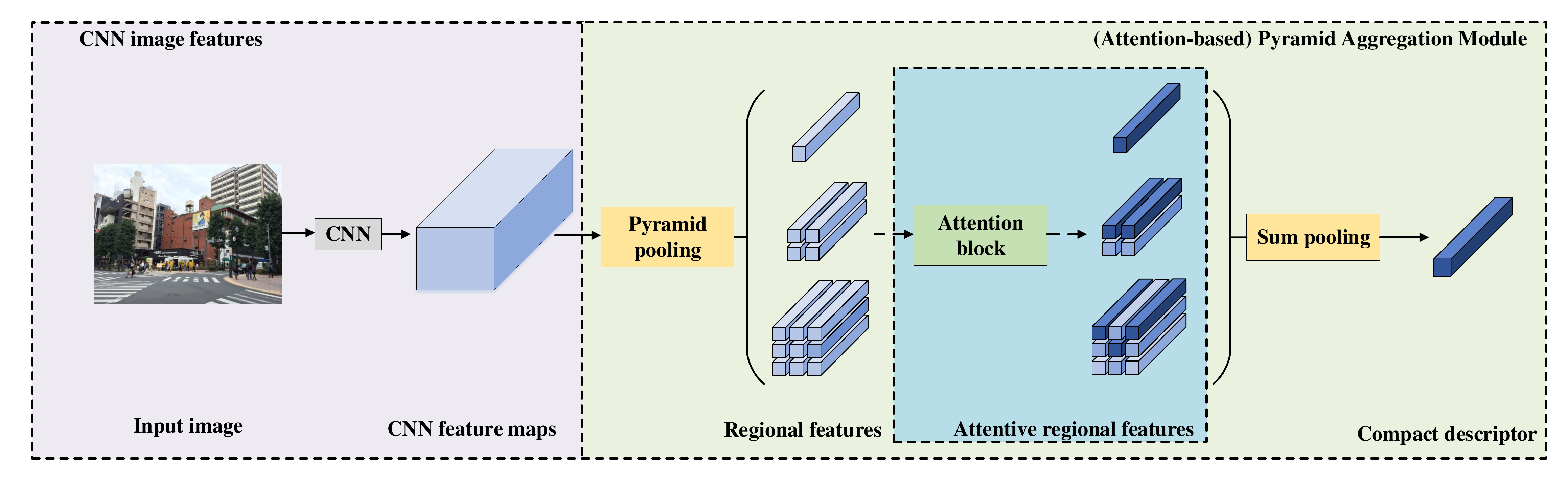}
		}
		\caption{Overview of the proposed APANet. For visualization, the pyramid pooling above adopts the spatial grids with scale (1,2,3). In our practice, we adopt a finer scales, \emph{e.g.,} scale (2,4,6,8). Same as conventional practice, the final compact descriptor are  $L_2$-normalized by default.}
		\label{SPA}
	\end{figure*}
	As a region-based aggregation method, our APANet is close to the works of 
	R-MAC \cite{tolias2015particular} and Deep Image Retrieval (DIR) \cite{gordo2016deep} which learns  R-MAC representation in a cleaned landmark dataset. We consider DIR as the baseline method and our APANet differs from it in the following aspects. First, we do not normalize or whiten the regional features before sum-aggregation. In our practice, these two operations perform well for the pre-trained CNNs but are unfavorable for CNN representation learning. Second, we have a finer scale choice for the spatial grids in APANet and increase the scale number to four, which helps the spatial grids prone to align with all the  buildings. 

    The attention block in our APANet helps to weight the regional features with regard to the distinctiveness and 
	 there are similar practices in CNN-based image retrieval tasks \cite{kalantidis2016cross,jimenez2017class,Hoang2017select,noh2017largescale}. Kalantidis \emph{et al.} \cite{kalantidis2016cross} improves sum pooling by exploiting the spatial weight and channel weight for each location on the feature maps. In \cite{jimenez2017class,mohedano2017saliency}, novel saliency priors are introduced for aggregating local CNN features. 
	However, to our knowledge, the effectivenesses of these weighting strategies have not been demonstrated for fine-tuned CNNs, or for evaluating the regional features.  In contrast to these methods, the attention block 
	is involved in the CNN architecture and optimized in the representation learning process, which enables an optimal weighting mechanism for aggregating regional features.



	\section{Attention-based Pyramid Aggregation Network}
	
	In this section, we describe the proposed attention-based pyramid aggregation network. We first introduce our pyramid aggregation module and show the modifications on pyramid pooling block for place recognition (Section \ref{sec:pyramid_pooling}). Then we depict the attention blocks adopted for weighting the regional features (Section \ref{sec:attention}). Section \ref{sec:street} presents the training objective.
	
	\subsection{Pyramid Aggregation Module} \label{sec:pyramid_pooling}
	Our pyramid aggregation (PA) module is composed of spatial pyramid pooling and sum pooling operations.
	Spatial pyramid pooling \cite{grauman2005pyramid,lazebnik2006beyond} was first introduced to CNN by \cite{he2014spatial}  to meet the fixed-length requirement of the fully-connected layer for visual recognition. Recently, it benefits scene parsing \cite{zhao2017pyramid} and saliency detection \cite{wang2017stagewise} tasks by encoding the contextual information.
	In PA module, spatial pyramid pooling acts as a region-based pooling method that helps encode the multi-size buildings in an image. To get a compact and discriminative descriptor, the PA module has modifications on the conventional pyramid pooling \cite{he2014spatial,zhao2017pyramid} in two aspects: (i) overlapping max pooling is utilized in the spatial grids so that they can better align with all the buildings, and (ii) all the regional features are sum-aggregated into a compact global descriptor rather than being concatenated. 
	
	
	As shown in Fig.~\ref{SPA},  given the feature maps from the CNN's last convolutional layer, we leverage pyramid pooling on them to aggregate the multi-scale regions. 
	Considering the convolutional feature maps of size $W \times H  \times D$, where $W \times H$ is the spatial resolution and $D$ is the number of channels, 
	the pyramid pooling has pooling window size in proportion to the size of feature maps. For a spatial grid with scale n,
	the output is fixed to $n \times n \times D$ and the size of pooling window is  $\left [  \left \lceil 2 \times W/(n+1) \right \rceil, \left \lceil 2 \times H/(n+1) \right \rceil \right ]$, stride is $\left [  \left \lceil W/(n+1) \right \rceil, \left \lceil H/(n+1) \right \rceil \right ]$ to enable approximately $50\%$ overlapping on each side, where $\lceil . \rceil$ is the ceiling operation. 
	 Then the regional features are arranged by scale as follow:
	\begin{equation}
	\mathbf{F_{R}} = \left [ \mathbf{f_{r,1}...f_{r,s_{1}^2}...f_{r,N}} \right ], \quad N=\sum_{j=1}^{n}s_{j}^2,
	\end{equation}
	where $\mathbf{f_{r,i}}$ is the $i^{th}$ regional feature with a size of $1*1*D$, $N$  the total number of regions, $S_{j}$ the size of $j^{th}$ scale, and there are $n$ scales in total. Thus we obtain a regional feature set $\mathbf{F_{R}}$  of size $1 \times N \times D$ and feed it to the sum pooling block to get the $1 \times 1 \times D$ global descriptor $\mathbf{F_{0}}$,
	\begin{equation}
\label{equ:F0} 
\mathbf{F_{0}} = \sum_{i=1}^{N}  \mathbf{f_{r,i}}.
\end{equation}

	

	\subsection{Attention Block}\label{sec:attention}
	
	In the pyramid aggregation module, all the regional features are sum-aggregated into a global descriptor. A critical problem exists that not all the regional features describe the regions of interest. The receptive fields of some regional features may be centered on the background or confusing objects like cars and trees in the urban environment. When sum-aggregation, these confusing regional features should be assigned lower weights to suppress their contributions. More precisely, we aim to assign each regional feature a score according to the distinctiveness. 
	Inspired by the attention mechanism popular in fine-grained recognition \cite{xiao2015application,Yan2017Fine} and video face recognition \cite{yang2016neural} tasks, we adopt the attention block to evaluate the regional features in PA module. 
	
	
	\subsubsection{Single Attention Block}
\begin{figure}[t]
		\setlength{\abovecaptionskip}{0.cm}
		\setlength{\belowcaptionskip}{-0.cm}
			\subfigure[Single attention block]{ \label{single attention}     
				\includegraphics[scale = 0.45,trim = 40 10 10 5,clip]{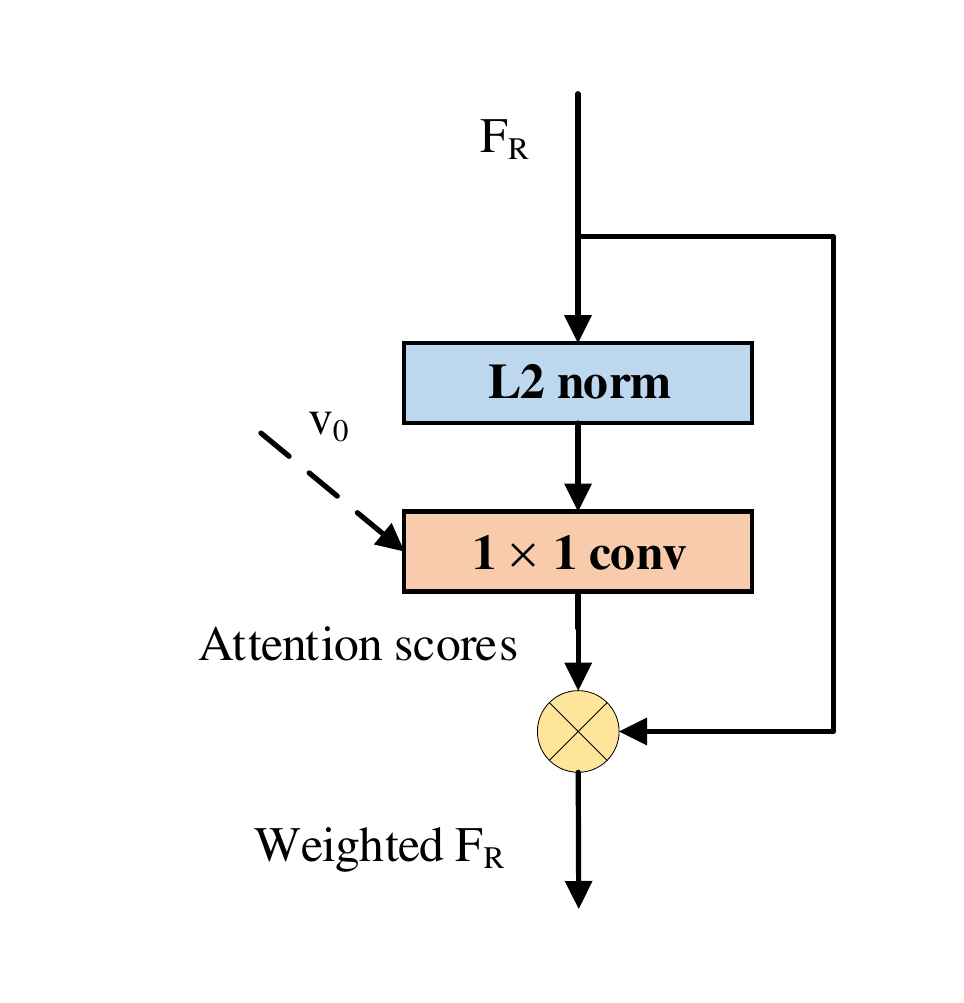}
			}
			\subfigure[Cascaded attention block]{\label{cascaded attention}   
				\includegraphics[scale = 0.45,trim = 45 10 10 5,clip]{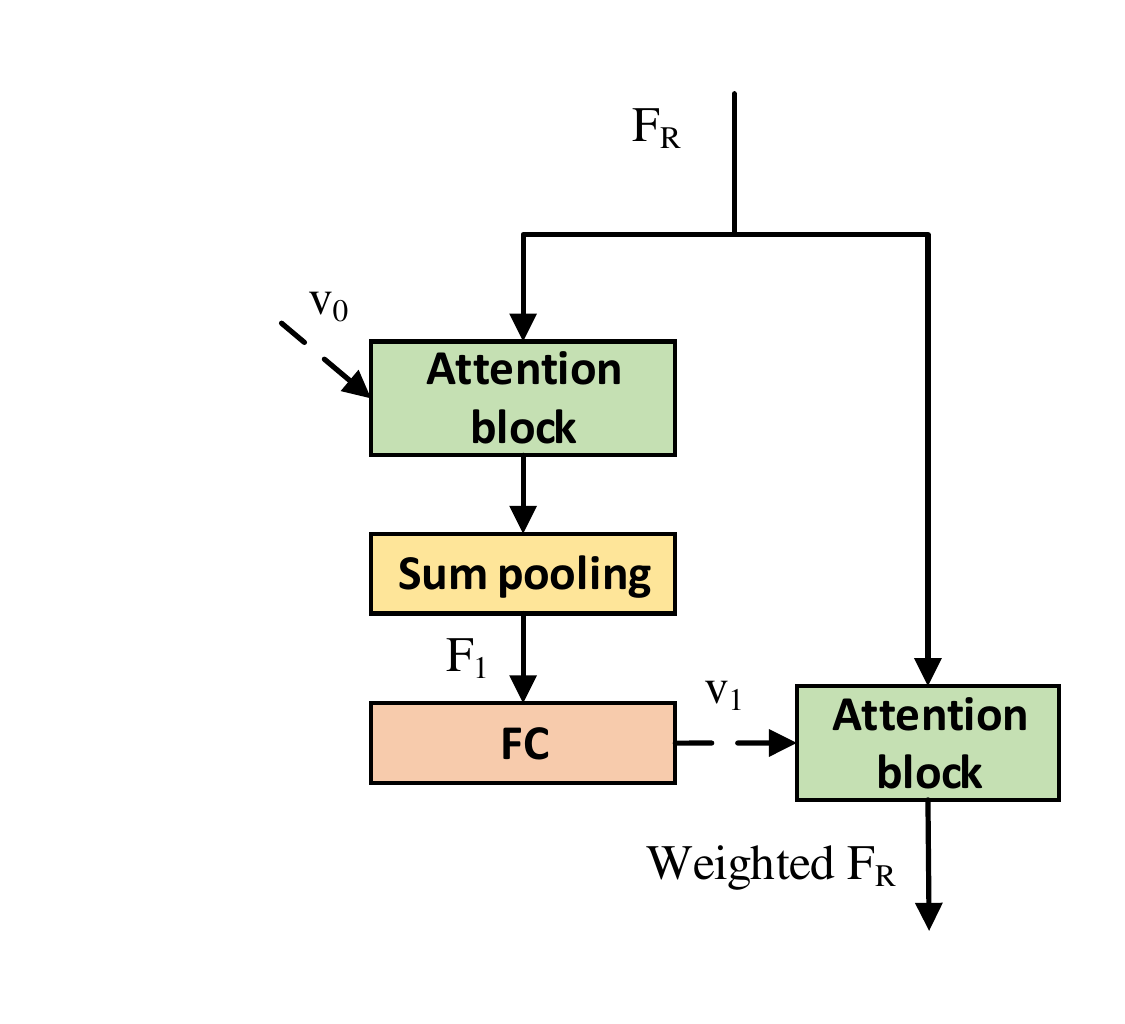}
			}		
		\caption{Schema of single attention block and cascaded attention block. $\otimes$ denotes element-wise multiplication operation.}
		\label{CSW}
	\end{figure}

	Our single attention block utilizes an $1\times1\times D$ convolutional layer on the regional features $\mathbf{F_{R}}$ for evaluating their distinctiveness (Fig. \ref{single attention}). The $D$-dimensional evaluation vector $\mathbf{v_{0}}$ (parameters of the $1\times1$ convolutional layer) inner-products all the regional features and produces attention scores, which will multiply the corresponding regional features and get weighted regional features. In this manner, the global descriptor $F_{1}$ after sum pooling can be expressed as following: 
	\begin{equation}
	\label{equ:F1} 
	\mathbf{F_{1}} = \sum_{i=1}^{N} a_{i} \mathbf{f_{r,i}}, \quad a_{i} = (\mathbf{v_{0}}*\mathbf{fn_{r,i}}),
	\end{equation}
   where $a_{i}$ is the attention score corresponding to the $i^{th}$ regional feature, $\mathbf{fn_{r,i}}$ is the $i^{th}$ regional feature after $L_2$-normalization, and $*$ denotes the inner product operation. We normalize the regional features when computing the attention scores because this helps the attention scores  be within a reasonable range, and we do not need to adopt an activation function on them. 

\subsubsection{Cascaded Attention Block}
	A single attention block works as an  evaluator for the regional features with regard to their distinctiveness. What if this evaluator has a content prior from the whole image? That is, the evaluator looks through the whole image first and then evaluates each region. We envision that this allows the evaluator to produce more reasonable scores for sum-aggregation. 
	To this end, we borrow the idea of the cascaded attention block which is used for video face recognition \cite{yang2016neural} and adjust it for weighting the regional features.
	The cascaded attention block incorporates two attention blocks as shown in Fig.~\ref{cascaded attention}. The first attention block is the same as the scheme of Fig.~\ref{single attention}, that produces attention scores for the regional features. 
	Having the global descriptor $\mathbf{F_{1}}$ in Eq.~(\ref{equ:F1}) after sum pooling, we impose a linear transformation on it by a fully-connected layer followed by hyperbolic tangent nonlinear activation function (tanh).
	The $D$-dimensional output $\mathbf{v_{1}}$ is used as a new evaluation vector for the second attention block, 
	\begin{equation}
	\mathbf{v_{1}} = \tanh (\mathbf{WF_{1}}),
	\end{equation}	
	where $\mathbf{W}$ and $\mathbf{b}$ are the parameters of the fully-connected layer. Compared to  $\mathbf{v_{0}}$ which is randomly initialized in the single attention block, the new evaluation vector $\mathbf{v_{1}}$ has a content prior from the global image descriptor $\mathbf{F_{1}}$.  
	
	By incorporating the attention block to the PA module and optimizing it with representation learning process, it can automatically learn to focus on the most discriminative regional features for place recognition.
	The effectiveness of the proposed attention block will be shown in Section~\ref{sec:apa}.
	\subsection{Learning Discriminative Representation with Triplet Ranking Loss}\label{sec:street}
	The PA module and the attention block are composed of differentiable conventional CNN operations, thus the proposed APANet is an end-to-end trainable model.
	Following NetVLAD, we fine-tune APANet with the weakly supervised triplet ranking loss on the Street View training datasets. Triplet loss has shown competitive effectiveness in deep metric leaning \cite{hoffer2015deep,wang2014learning}, face identification \cite{schroff2015facenet} and image retrieval tasks \cite{arandjelovic2016netvlad,gordo2016deep}. Learning to rank the positive and negative images in the triplets enables the network to produce discriminative descriptors. Detailed descriptions about the adopted weakly supervised triplet ranking loss and strategy for mining the training tuples can be found in \cite{arandjelovic2016netvlad}.
	
	
	

	\section{PCA Power Whitening}
	\begin{figure*}[t]
		\setlength{\abovecaptionskip}{0.cm}
		\setlength{\belowcaptionskip}{-0.cm}
		\centerline{
			\subfigure[PCA rotation ($\mathbf{X_{R}}$)]{ \label{fig:PCAa}     
				\includegraphics[scale = 0.15]{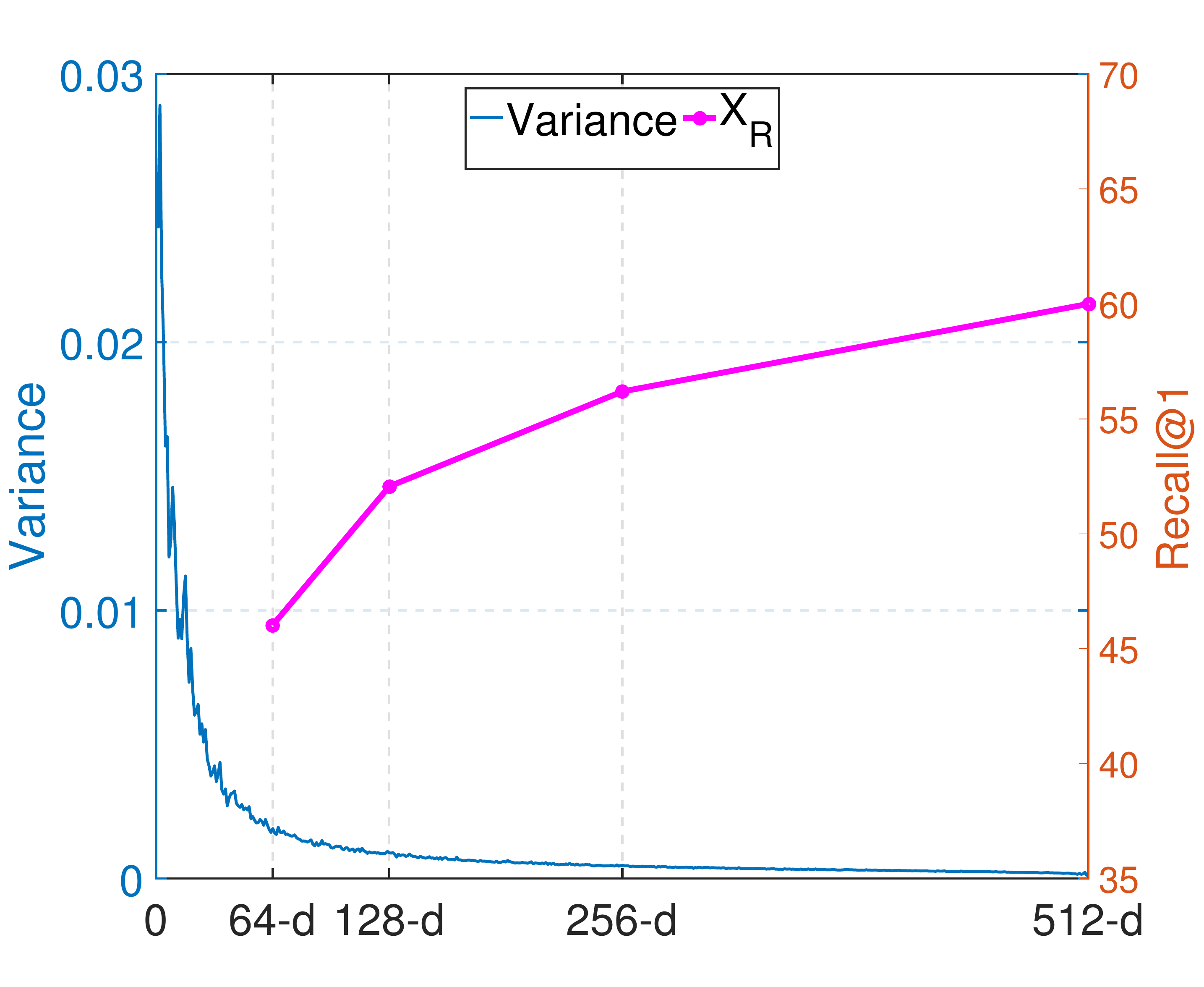}
			}
			\subfigure[PCA whitening ($\mathbf{X_{W}}$)]{ \label{fig:PCAb}     
				\includegraphics[scale = 0.15,trim = 20 0 5 0,clip]{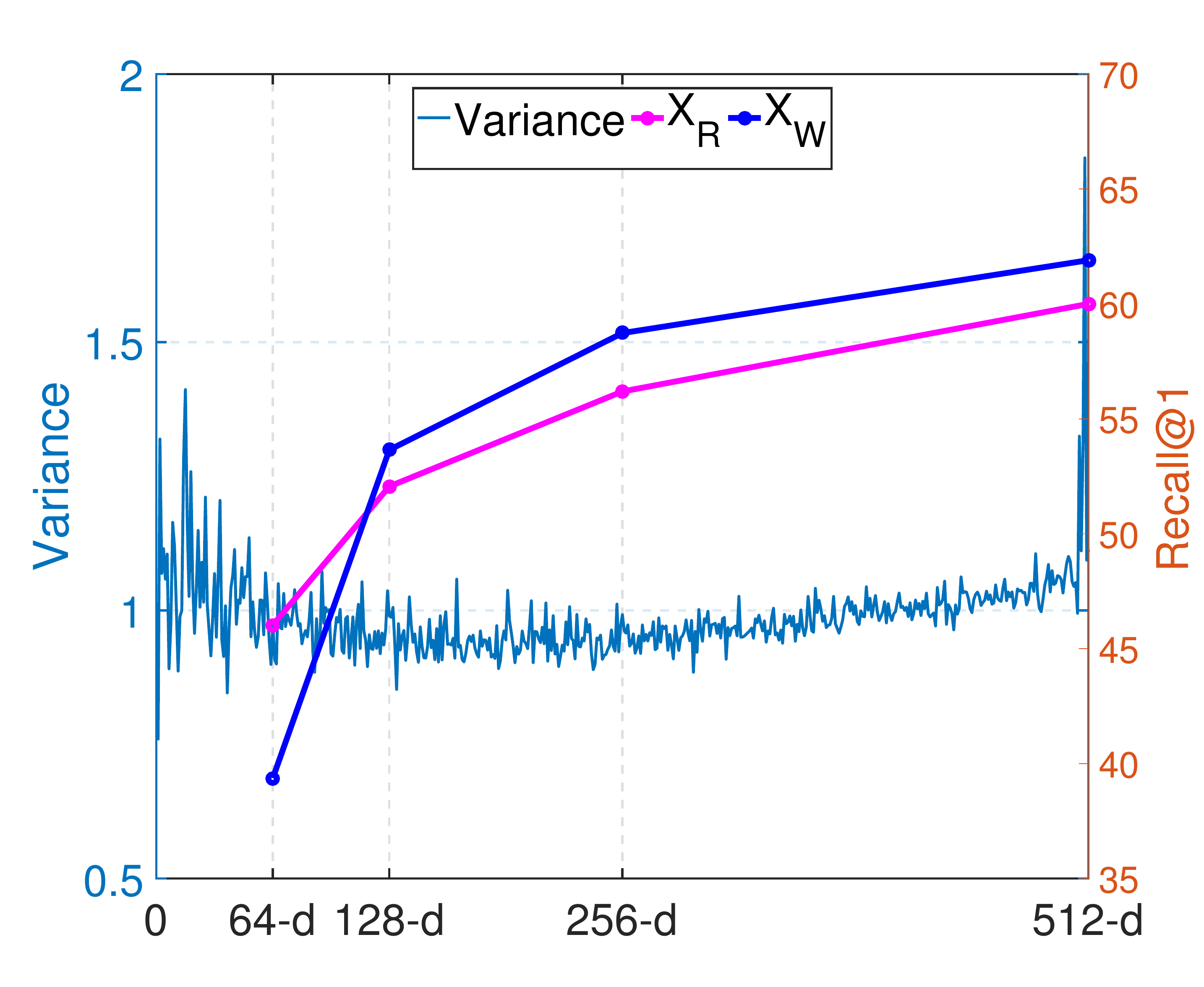}
			}
			\subfigure[Manually scaling ($\mathbf{X_{M}}$)]{\label{fig:PCAc}   
				\includegraphics[scale = 0.15,trim = 20 0 5 0,clip]{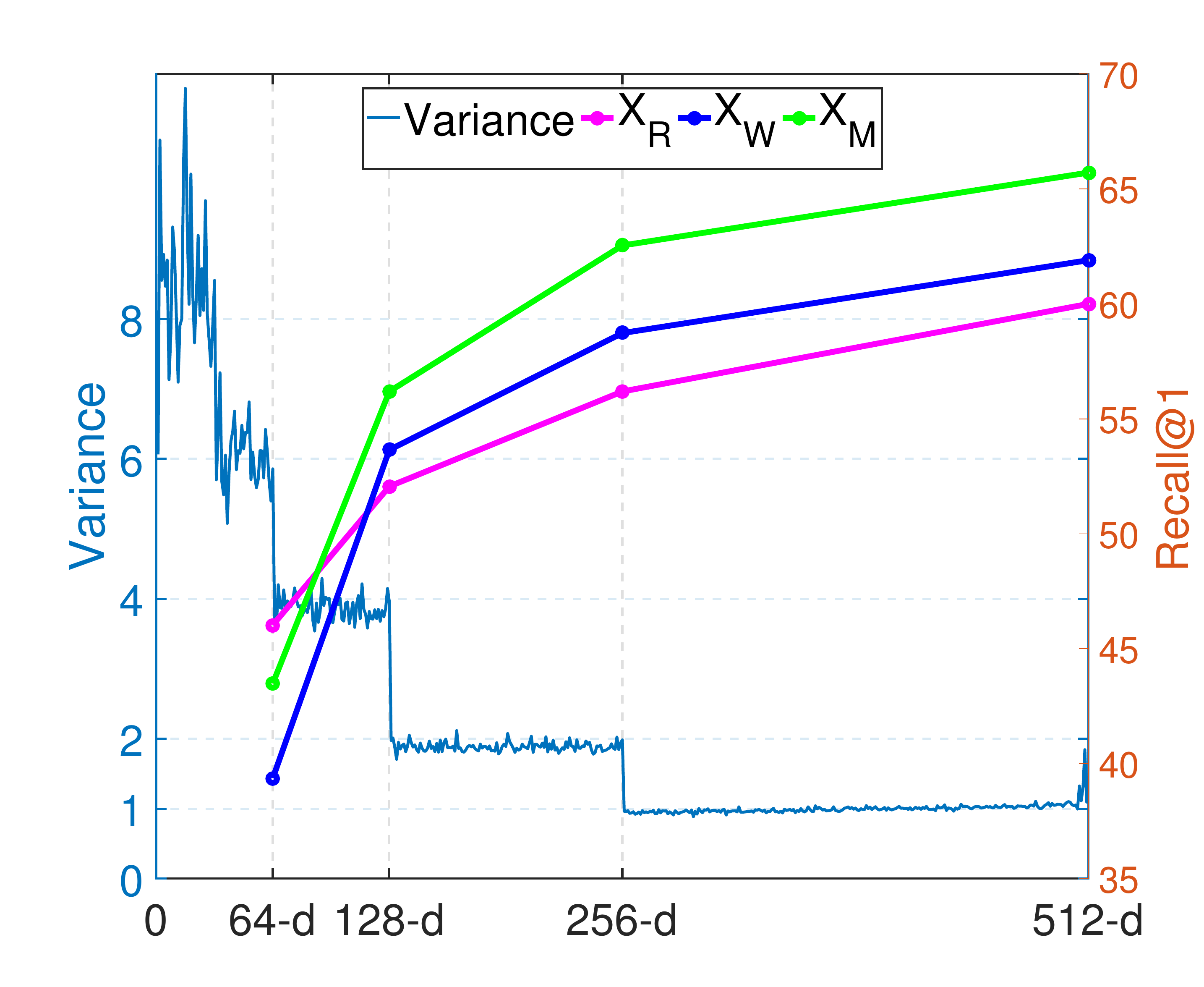}
			}
			\subfigure[PCA power whitening ($\mathbf{X_{PW}}$)]{\label{fig:PCAd}   
				\includegraphics[scale = 0.15,trim = 10 0 5 0,clip]{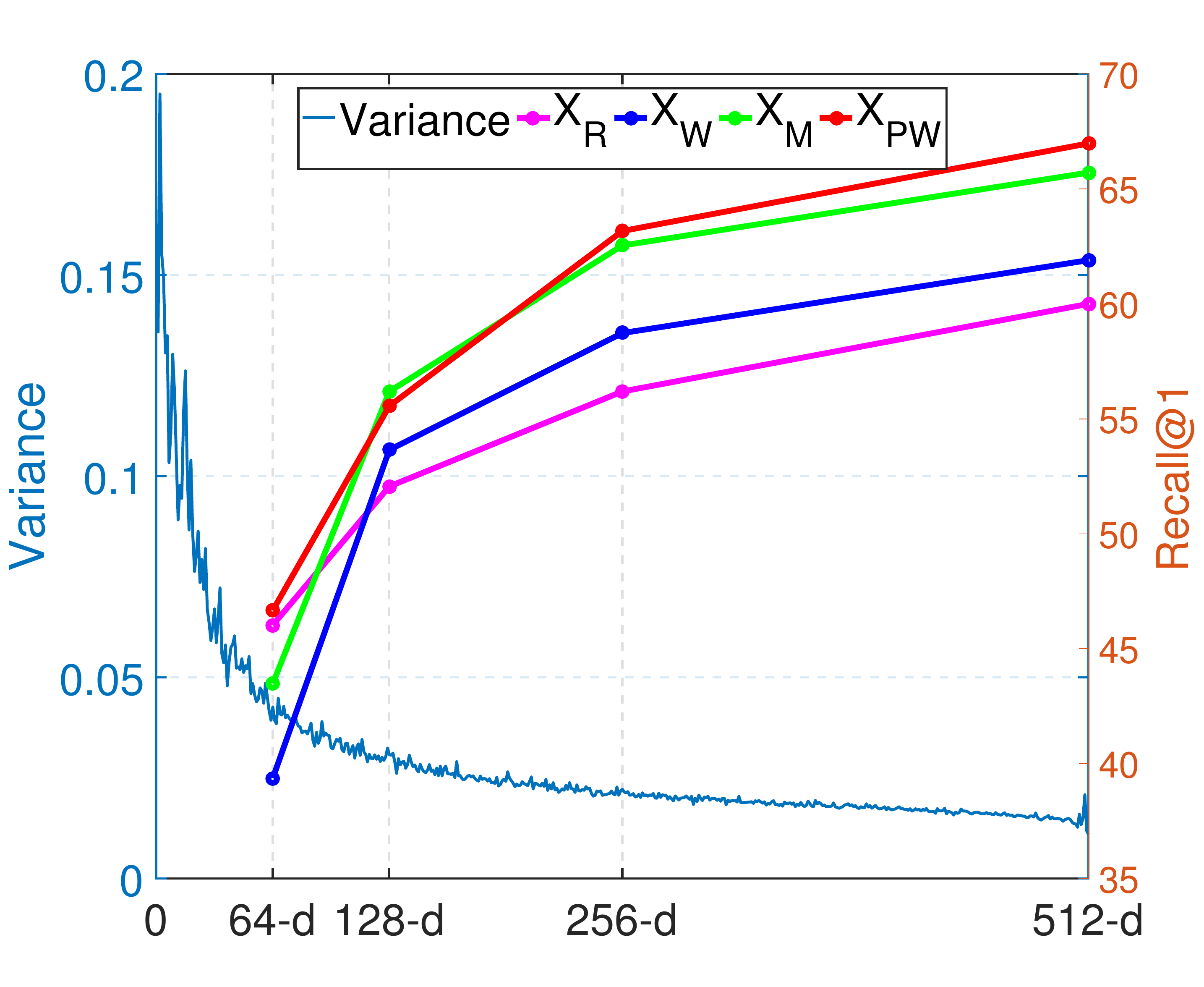}
				\label{PCA-rootc}
			}
		}
		\caption{Variance versus place recognition accuracy. For visualization, variances are calculated from the 512-D APANet representations on Tokyo 24/7 database images before normalization while the recalls are from normalized representations.}
		\label{PCA-root}
	\end{figure*}
	\textbf{PCA Whitening.} Whitening is an effective post-processing approach in image retrieval. It helps to solve the problem of over-counting and co-occurrences \cite{jegou2012negative}, thereby improving the retrieval accuracy for a series of works \cite{tolias2015particular,babenko2015aggregating,arandjelovic2016netvlad}. The whitening parameters are usually learned by PCA on an independent dataset. Let us consider a PCA rotation on the descriptors $\mathbf{X}$, 
	\begin{equation}
	\mathbf{X_{R}} = \mathbf{P}^{T}\mathbf{X},
	\end{equation}
	where $\mathbf{P}$ is the $D \times D$ PCA rotation matrix, X is  $L_2$-normalized and optionally zero-centered. 
	After  PCA rotation, the first few dimensions of $\mathbf{X_{R}}$ preserve most energy and have larger variances than the later dimensions as shown in Fig.~\ref{fig:PCAa}. Meanwhile, the over-counting in the representations (where $\mathbf{P}$ was learned) is captured in $\mathbf{P}$ and mostly influences the first few dimensions of $\mathbf{X_{R}}$ \footnote{A detailed description can be seen in Section 4 of \cite{jegou2012negative}.}.
	Whitening operation (Equation~\ref{pcaw}) address over-counting problem by balancing the energy (\emph{i.e.} variance) of each dimension in $\mathbf{X_{R}}$. As illustrated in Fig.~\ref{fig:PCAb}, whitening  scales each dimension of $\mathbf{X_{R}}$ to unit variance, which has a pretty enhancement on the performance.
	\begin{equation}
	\label{pcaw}
	\mathbf{X_{W}} = diag(\lambda_{1}^{-\frac{1}{2}},...,\lambda_{D}^{-\frac{1}{2}}) \mathbf{P}^{T}\mathbf{X},
	\end{equation}
	where $\lambda_{i} $ is the eigenvalue associated with $i^{th}$ eigenvector in $\mathbf{P}$.
	
\textbf{Our observation.} However, we argue that PCA whitening may excessively penalize the problem of over-counting. In fact, it is beneficial that the variance of the former dimensions are somehow preserved, so that a balance is achieved between reducing over-counting and preserving the energy distribution of the features. On the basis of $\mathbf{X_{W}}$, we manually increase the variances of the first 256 dimensions with different multiples and the performance improvement over $\mathbf{X_{W}}$ can be seen in Fig.~\ref{fig:PCAc}. 

\textbf{PCA power whitening.} Motivated by the observation above, we propose a PCA power whitening (PCA-pw) strategy, which provides more reasonable variance scaling based on the eigenvalue $\lambda_{i} $, 
	\begin{equation}
	\label{rooteq}
	\mathbf{X_{PW}} = diag(\lambda_{1}^{-\frac{1}{2}\times\alpha },...,\lambda_{D}^{-\frac{1}{2}\times\alpha}) \mathbf{P}^{T}\mathbf{X},
	\end{equation}
	where $0 \leq \alpha \leq 1$ is the parameter of scaling. 
	Usually 0.5 is a reasonable value for $\alpha$ and we choose it by default for PCA-pw. The performance of PCA-pw is presented in Fig~\ref{fig:PCAd}. We can clearly observe that PCA-pw enables largest performance improvement compared to the all the methods above. 
	
	PCA rotation and PCA whitening can be seen as special cases of PCA-pw  where $\alpha = 0$ and 1, respectively. So the PCA-pw can reasonably limit the impact of over-counting by setting a proper value of $\alpha$, and providing a maximum level of improvement on the retrieval performance. This argument will be further supported by more experiment results in section~\ref{sec:power ex}.
	In our practice, PCA-pw improves a series of retrieval baselines \cite{arandjelovic2016netvlad,tolias2015particular,babenko2015aggregating} which use PCA whitening.


	\section{Experiment}\label{sec:exp}
	In this section, we present the place recognition datasets, analyze the scale choice for pyramid pooling block and demonstrate effectiveness of the attention block. Then we apply PCA power whitening on four representative aggregation methods. Finally, we compare APANet  to the state-of-the-art and show its generalization ability on the standard image retrieval datasets. 
	
	\subsection{Datasets and Implementation Details} \label{sec:datasets}
	We evaluate the proposed APANet on two place recognition datasets: \textbf{Pitts250k-test} the \cite{torii2013visual} and \textbf{Tokyo 24/7} \cite{torii201524} datasets. Pitts250k contains 254k perspective images generated from 10.6k Google Street View panoramas in Pittsburgh area. Pitts250k-test is a subset of Pitts250k and has around 83k database images, 8k query images. Tokyo 24/7 is a challenging dataset that contains 76k database images and 315 query images captured by different mobile phones cameras at daytime, sunset and night. 
	The Street View training datasets \cite{arandjelovic2016netvlad} consist of Pitts30k-train and Tokyo Time Machine (TokyoTM) dataset.  We choose the Pitts30k-train or TokyoTM dataset  for fine-tuning according to the testing dataset. 
	
    \textbf{Evaluation metric.}
	For these two evaluation datasets, we follow the standard evaluation protocol in \cite{arandjelovic2016netvlad,torii2013visual,torii201524}. The performance is measured by the percentage of correctly recognized queries at given top $N$ candidates (Recall@N). A query image is deemed to be correctly recognized if at least one of the top $N$ candidate database images are within 25 meters from the ground truth position.  
	
	\begin{table}[t]
		\caption{Performance of R-MAC and PANet with different scale choices and SNW operations. PANet (1234) has four scales spatial grids  and 30 regional features ($1, 2 \times 2, 3 \times 3, 4 \times 4$). The best results are highlighted in \textbf{bold}. All these methods are based on AlexNet.}
		\label{tab:scale}
		\centerline{
			\begin{tabular}{|l|l|c c c|}
				\hline
				\multirow{2}*{Method}  & \multirow{2}*{Regions}&\multicolumn{3}{c|} {Pitts250k-test} \\
				\cline{3-5}
				&& R@1& R@5& R@10 \\
				\hline
				R-MAC w/o SNW&20&68.35 &82.93	&87.63	\\
				PANet (1234)&30& 68.86&83.47	&87.57	\\
				PANet (2468)&120& \textbf{69.69}&	\textbf{83.77}&	87.85\\
				PANet (2345678)&203&69.24&83.19	&87.23	\\
				\hline\hline
				R-MAC \cite{gordo2016deep}&20& 68.24&83.70&	\textbf{87.97}\\
				PANet (1234) + SNW&30& 67.27&82.56	&86.70	\\
				PANet (2468) + SNW&120& 63.60&79.78&84.65\\
				PANet (2345678) + SNW&203&67.21&82.11	&86.35	\\
				\hline
			\end{tabular}
		}
	\end{table}

\begin{table*}[ht]
	\caption{Comparison of two aggregation methods when the single or cascaded attention block is integrated. All these methods are based on AlexNet. The best results of each method are highlighted in \textbf{bold}.}
	\label{tab:spatial weight}
	\setlength{\abovecaptionskip}{0.cm}
	\setlength{\belowcaptionskip}{-0.cm}
	\centerline{
		\begin{tabular}{|l|c|c|c c c|c c c|}
			\hline
			\multirow{2}*{Method} & \multirow{2}*{ Single }&\multirow{2}*{ Cascaded } & \multicolumn{3}{c|} { Pitts250k-test } &\multicolumn{3}{c|} { Tokyo 24/7 }\\
			\cline{4-9}
			&&& Recall@1& Recall@5& Recall@10& Recall@1& Recall@5& Recall@10 \\
			\hline	
			\multirow{3}*{Sum pooling}& &&58.53&75.56&82.08&28.57&42.22&53.02\\
			&${\surd}$&&60.76&77.23&82.62&28.25&\textbf{46.35}&53.33\\
			&&${\surd}$&\textbf{63.84}&	\textbf{78.95}&	\textbf{84.12}&\textbf{29.84}&46.03&\textbf{54.60}\\
			\hline
			PANet  &&&69.69&83.77&87.85&33.65&48.57&53.02\\
			\hline
			\multirow{2}*{APANet} &${\surd}$&&71.20&85.75&89.41&34.92&49.21&53.65\\
			&&${\surd}$&\textbf{72.69}&	\textbf{86.38}&\textbf{89.92}&\textbf{38.41}&\textbf{53.97}&\textbf{61.27}\\
			\hline
		\end{tabular}
	}
\end{table*}
\begin{figure*}	[ht]
	\setlength{\abovecaptionskip}{0.cm}
	\setlength{\belowcaptionskip}{-0.cm}
	\centerline{
		\includegraphics[scale= 0.8,trim = 5 20 20 20,clip]{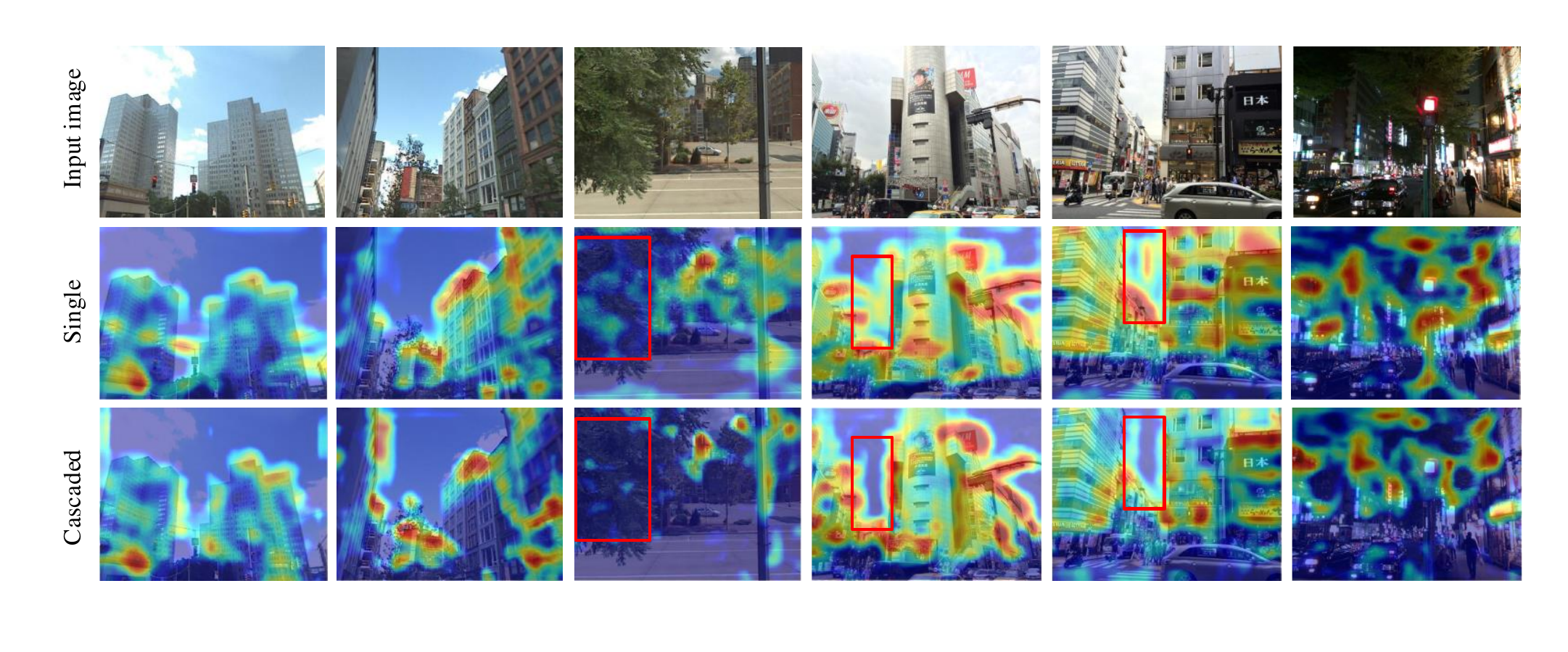}
	}
	\caption{Visualization of attention score maps in heat maps. (top) input images. (middle) attention scores from a single attention block defined in Fig. \ref{single attention}. (bottom) attention scores from the cascaded attention block defined in Fig. \ref{cascaded attention}.}
	\label{heat}
\end{figure*}
	\textbf{Implementation details.}
	The pre-trained AlexNet \cite{krizhevsky2012imagenet} and VGG-16 \cite{simonyan2014very} architectures are adopted as the base CNN architectures for fine-tuning and both are cropped at the last convolutional layer before Relu. Besides the attention-based pyramid aggregation module, the  R-MAC and sum pooling are also adopted as the aggregation layer for fine-tuning. For all these methods, we use margin $m=0.1$, batch size of 4 tuples, SGD with 
	initial learning rate $l_{0}$ of 0.001 for Pitts30k-train and 0.0005 for the TokyoTM dataset, and an exponential decay $l _{0}\exp(-0.1(i-1))$ over epoch $i$, momentum 0.9, weight decay 0.001. We use Xavier initialization \cite{glorot2010understanding} for the attention blocks, whose learning rate is ten times of the formal convolutional layers. When testing, the whitening parameters are learned from 10k images randomly sampled from the Pitts30k-train or TokyoTM dataset, the same as NetVLAD. For fair competition with NetVLAD, we do not perform data augmentation or three-clip testing as \cite{kim2017learned} did.
	
	\subsection{Evaluation of APANet} \label{sec:apa}
	\subsubsection{Scale Choice and Baseline.}
	We consider R-MAC as the baseline method and fine-tune the base architecture with R-MAC on the Street View training datasets. The network configuration is the same as DIR \cite{gordo2016deep}. Our PA module differs from R-MAC by adopting a different region choice and removing the shift, normalization and whitening (SNW) operations on the regional features. 
	
	We first analyze the scale choice for our pyramid pooling block. 
	R-MAC has three scales of rigid grids which define around 20 regions $(1 \times 2, 2 \times 3, 3 \times 4)$. The grid scale of R-MAC is too coarse because the receptive field of each grid cell covers the whole image, which is ineffective for encoding the local cues in an image for complex scenes in place recognition. Hence, we adopt a finer scale choice in the proposed PA module (\emph{i.e.} PANet) and  increase the scale number to four. The upper part of Table~\ref{tab:scale} presents the comparisons of different region choices. We observe that ``PANet (2468)'' exceeds ``PANet (1234)'' and ``R-MAC w/o SNW'' by adopting a finer scale choice and more region numbers. This is because the likelihood that the regions of interest are well-aligned increases as the number of regions increases. However, increasing the number of regions  may also incur more confusing regions  to corrupt the image similarity measurement. It can be proven that ``PANet (2345678)'' with the largest number of scales and regions does not perform best. 
	
	Then we discuss the SNW operations in R-MAC and PANet. As shown in Table~\ref{tab:scale}, R-MAC performs similarly when attaching or removing SNW operations. But the performance of PANet with SNW operations drops significantly when the scales get finer. The interpretation is that the number of confusing regional features increases when the scale gets finer. These confusing regional features usually carry lower norms than the discriminative ones while SNW operations may highlight their contributions. Thus in the following experiments, we use four grid scales $(2, 4, 6, 8)$ for the pyramid pooling block by default and do not adopt SNW operations.
	
	
	

	\begin{table*}[ht]
		\caption{Comparison of PCA whitening and PCA power whitening (PCA-pw). All the results are from the 512-D representations based on VGG-16 architecture.} 
		\label{tab:spa aggregation}
		\setlength{\abovecaptionskip}{0.cm}
		\setlength{\belowcaptionskip}{-0.cm}
		\centerline{
			\begin{tabular}{|l|l|c c c|c c c|}
				\hline
				\multirow{2}*{Method} & \multirow{2}*{Whitening} & \multicolumn{3}{c|} {Pitts250k-test}&\multicolumn{3}{c|} {Tokyo 24/7} \\
				\cline{3-5} \cline{6-8}
				&& Recall@1& Recall@5& Recall@10 &Recall@1& Recall@5& Recall@10 \\
				\hline
				\multirow{3}*{Mac}&W/o whitening & 77.01&	88.73&	91.97&38.41&52.70&62.22\\
				&PCA whitening & 73.21&	86.03&	89.77&25.40&40.63&45.40\\
				&PCA-pw& \textbf{79.19}&	\textbf{90.12}&	\textbf{93.09}&35.56&52.06&60.95\\
				&PCA-pw $(\alpha=0.1)$& 78.25&89.44&92.26&\textbf{38.73}&\textbf{53.97}&\textbf{62.54}\\
				\hline
				\multirow{2}*{Sum pooling}&PCA whitening& 74.13 & 86.44&90.18&44.76&	60.95&	70.16\\
				&PCA-pw	& \textbf{75.63}&	\textbf{88.01}&	\textbf{91.75}&\textbf{52.70}&\textbf{67.30}&\textbf{73.02}\\
				\hline
				\multirow{2}*{NetVLAD \cite{arandjelovic2016netvlad}}&PCA whitening& 80.66 & 90.88&93.06&\color{blue}\textbf{60.00}&73.65&79.05\\
				&PCA-pw& \color{blue}\textbf{81.95}&	\color{blue}\textbf{91.65}&	\color{blue}\textbf{93.76}&58.73&\color{blue}\textbf{74.60}&\color{blue}\textbf{80.32}\\
				\hline
				\multirow{2}*{APANet}&PCA whitening& {82.32} & {90.92}&{93.79}&{61.90}&{77.78}&{80.95}\\
				&PCA-pw& {\color{red}\textbf{83.65}} & {\color{red}\textbf{92.56}}&{\color{red}\textbf{94.70}}&\color{red}\textbf{66.98}&\color{red}\textbf{80.95}&\color{red}\textbf{83.81}\\
				\hline	
			\end{tabular}
		}
	\end{table*}
	
	\subsubsection{Effect of Attention Block.} 
	Note that attention blocks can also weight the local CNN features for sum pooling method.
	We evaluate the attention blocks by employing them on the sum pooling and our PA module (namely APANet). Training is conducted on  the Street View training datasets and the results are displayed in Table~\ref{tab:spatial weight}. We observe that the attention blocks improve all these two aggregation methods on two datasets and the cascaded attention block always works better than the single one. In addition, the PA module has significantly better performances than sum pooling, which also indicates the effectiveness of PA module.
	
	For visualization, attention scores of sum pooling are presented in Fig.~\ref{heat}. These two attention blocks really work as we expected, \emph{i.e.,} focusing on the architectures and assigning lower attention scores to confusing objects such as pedestrians, cars and trees.
	Fig.~\ref{heat} also suggests that cascaded attention block has more localized attention score maps than the single one, thus paying more attention to the most discriminative regional features and resulting in more discriminative image representations.
	This observation can be viewed at column 3-5.  For example, the buildings are severely occluded by trees at column 3. In this case, the cascaded attention block can still focus on the buildings while in comparison the single attention block is less concentrated.     
    At columns 4 - 5, the background between two buildings is usually assigned lower scores in the cascaded attention score maps while it is considerably activated in single attention score maps. 
    In rest of the paper, we use cascaded attention block for our APANet unless otherwise specified.

	\subsection{PCA Power Whitening} \label{sec:power ex}
	To assess the effectiveness and universality of the proposed PCA power whitening (PCA-pw), we compare it with PCA whitening on four representative aggregation methods, \emph{i.e.} global max pooling (Mac), sum pooling, NetVLAD and our APANet. 
	We learn the image representations with these aggregation methods on the Street View training datasets and present the results in Table~\ref{tab:spa aggregation}. Several things can be observed. First, PCA-pw usually performs better than PCA whitening, especially on the Tokyo 24/7 dataset where the over-counting problem from the buildings is not so serious than Pitts250k-test. 
	Second, for the Mac representations, the problem of over-counting is not a big deal. 
	PCA Whitening even decreases the performances on both datasets because it excessively penalizes the over-counting. By alleviating these, PCA-pw improves the performance of Mac representations on Pitts250k-test dataset, but still decreases on Tokyo 24/7. We find there is a small enhancement on Tokyo 24/7 when setting the scaling factor $\alpha$ to 0.1, which reflects the degree of over-counting that Mac representations suffer on two datasets. Third, APANet representations perform best on both datasets, regardless of which whitening strategy is adopted.
	
	In summary, because of the aggregation method, dataset characteristics and the local CNN features themselves, the CNN-based image representations may suffer from over-counting problem more or less. The over-counting is solved by PCA Whitening in an extreme way while the proposed PCA-pw can better address it by setting a reasonable scaling factor $\alpha$, thereby providing consistently performance improvement for image retrieval. 

	\subsection{Comparison with State-of-the-Art} 
\begin{figure*}[t]
		
		\setlength{\abovecaptionskip}{0.cm}
		\setlength{\belowcaptionskip}{-0.cm}
		\centerline{
			\subfigure[\scriptsize Pitts250k-test]{
			\includegraphics[scale = 0.23,trim = 5 5 20 0,clip]{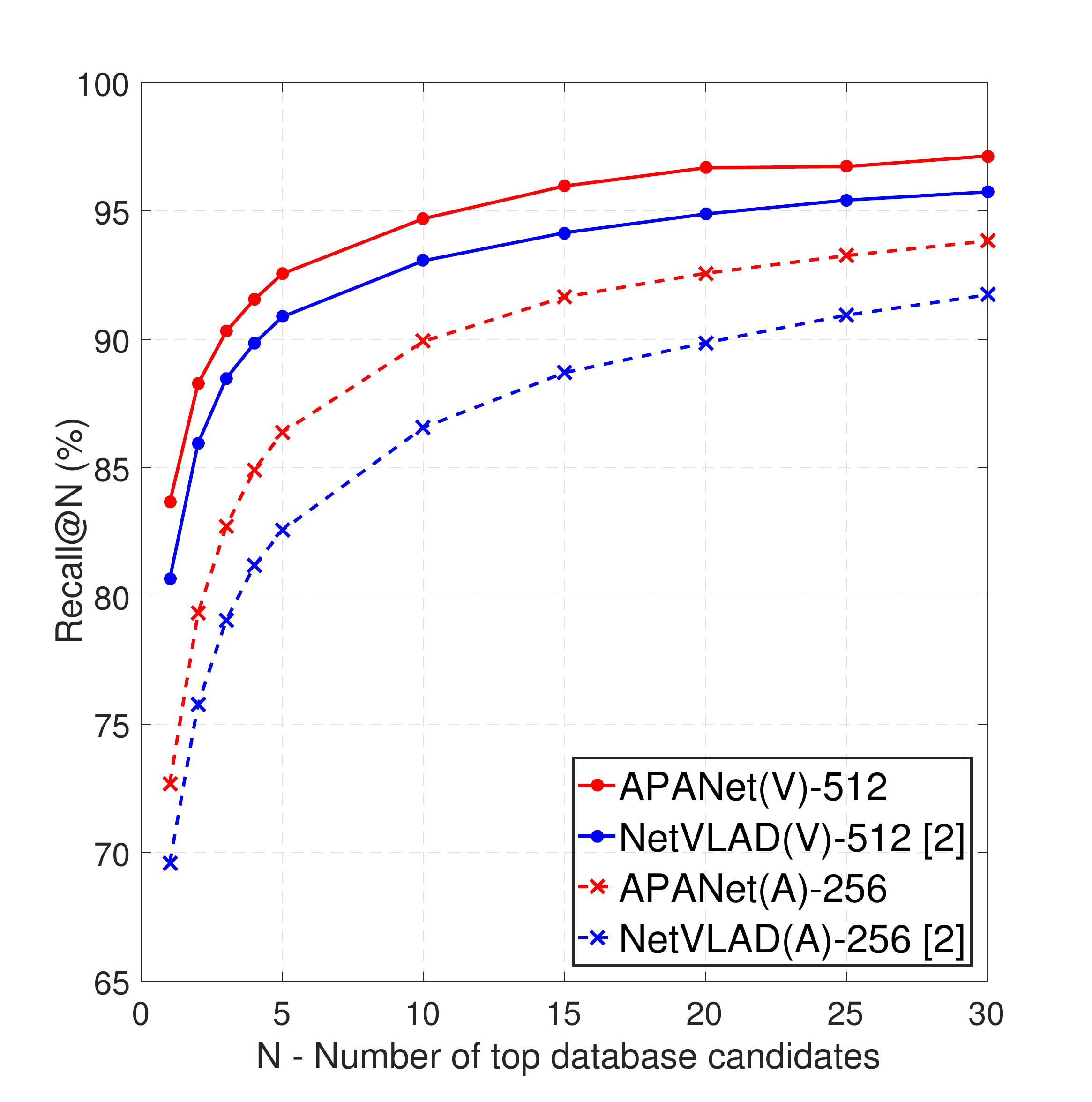}
			}
			\subfigure[\scriptsize Tokyo 24/7]{
			\includegraphics[scale = 0.23,trim = 5 5 20 0,clip]{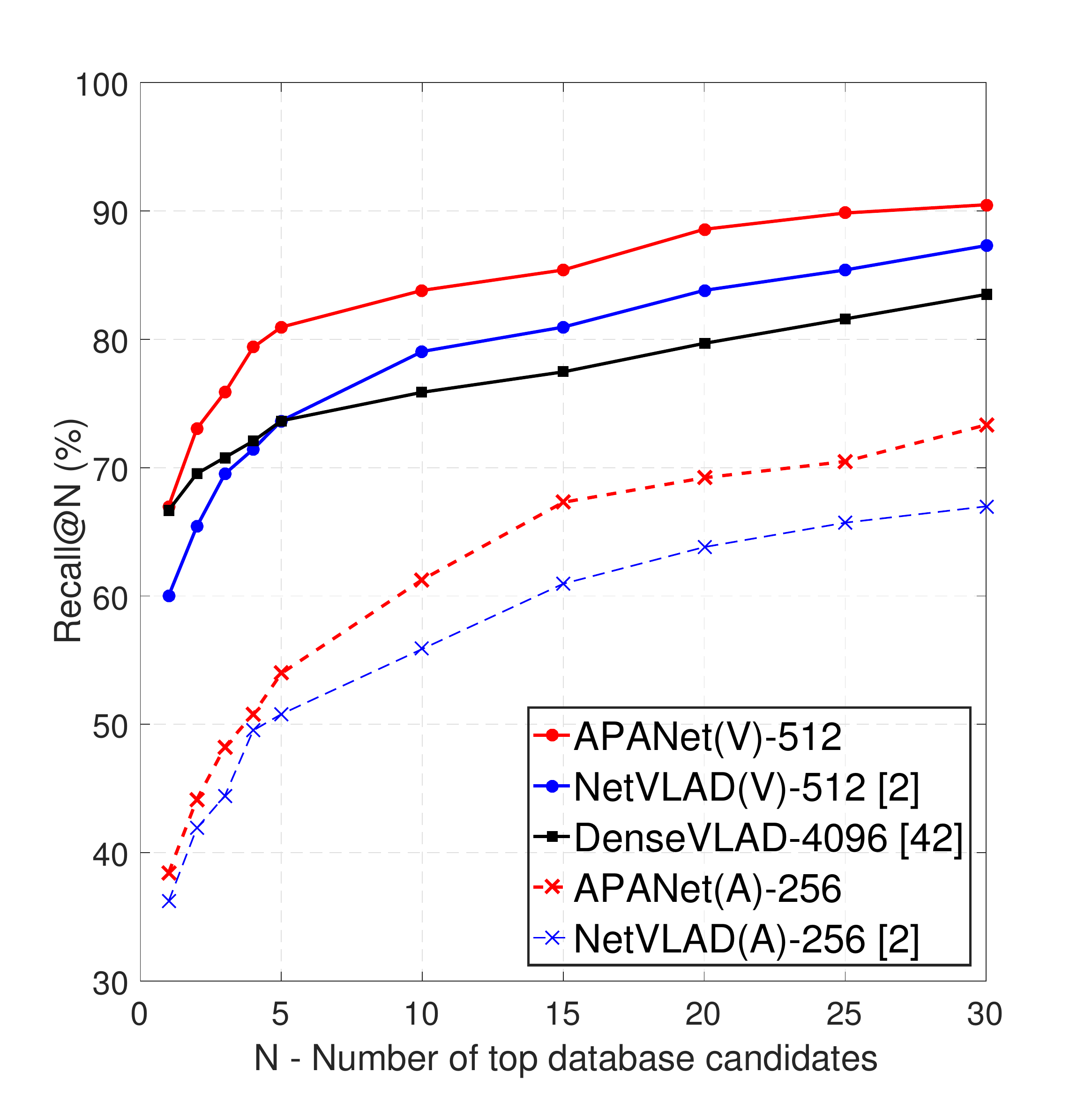}
			}
			\subfigure[\scriptsize Tokyo 24/7 sunset/night]{
			\includegraphics[scale = 0.23,trim = 5 5 20 0,clip]{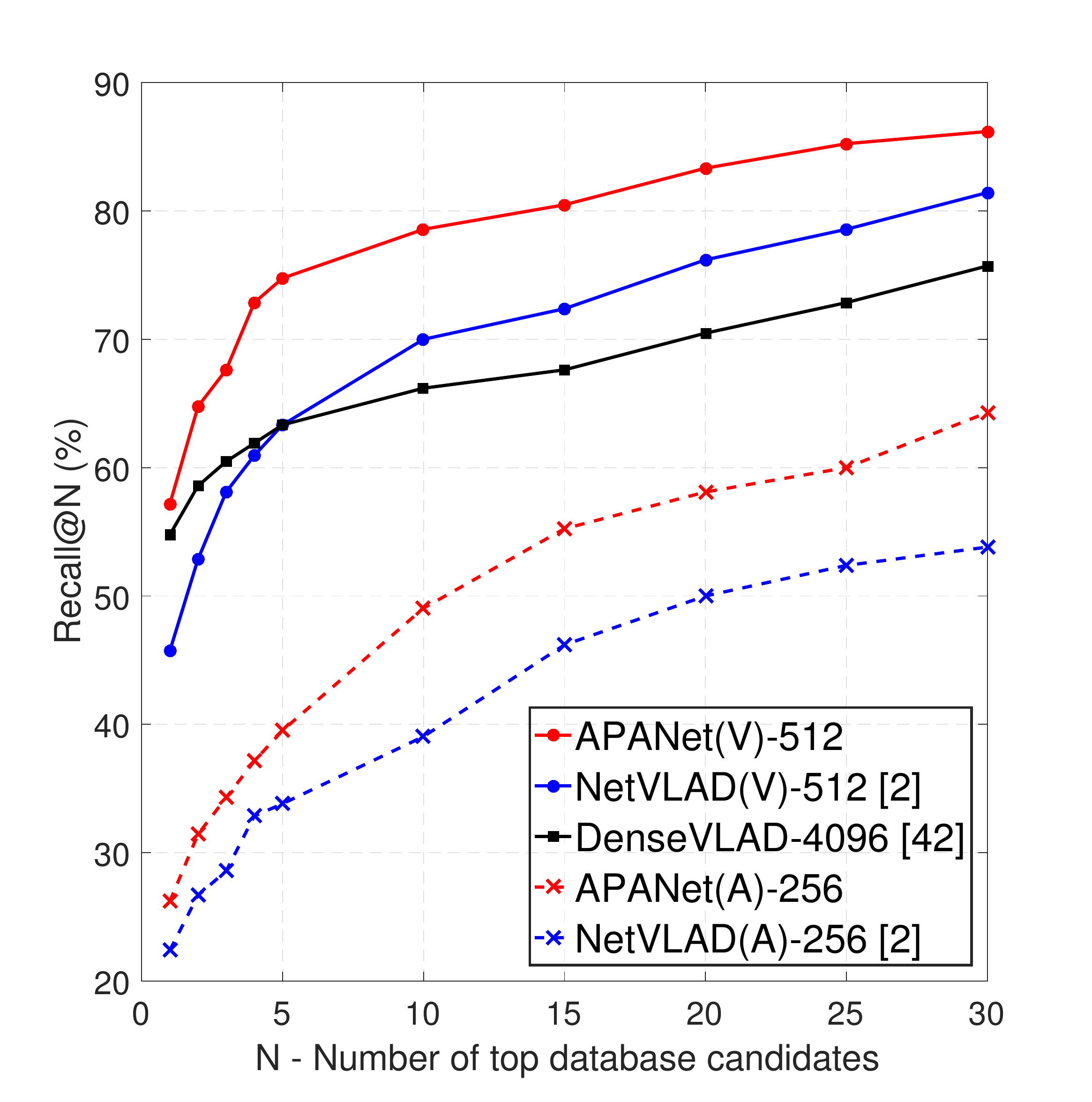}
			}
		}
		\caption{Comparision of recalls with previous state-of-the-arts. The base CNN architecture is denoted in brackets: (V) for VGG-16 and (A) for AlexNet model. The dimensionality is followed. }
		\label{PT}
	\end{figure*}

	The Dense-VLAD [12] combining view synthesis with densely sampled VLAD descriptors enables  recognition across the variations in viewpoint and illumination condition. And the NetVLAD-based deep representations \cite{arandjelovic2016netvlad,kim2017learned} achieve the state-of-the-art performance in place recognition datasets.
	We compare APANet with these methods and present the result in Fig.~\ref{PT} \footnote{We do not include the curves of  \cite{kim2017learned} in the figure because we can not get the recall curves or the trained models from the authors, and the performance of  \cite{kim2017learned}  is similar to NetVLAD on Pitts250k-test dataset.}. 
    It can be seen that APANet consistently outperforms NetVLAD using the AlexNet and VGG-16 architecture on all the datasets.  For VGG-16 architecture, the Recall@1 accuracies of APANet exceed NetVLAD with margins of $2.99\%$ and $6.98\%$  on Pitts250k-test and Tokyo 24/7 dataset, respectively. 
	Even if we didn't perform data augmentation on the lighting conditions, the gap between APANet(V) and the Dense-VLAD is even more pronounced on the challenging Tokyo 24/7 sunset/night subset, which demonstrates that the proposed APANet is robust to changes of illumination and viewpoint. 

	\begin{figure}
	\setlength{\abovecaptionskip}{0.cm}
	\setlength{\belowcaptionskip}{-0.cm}
	\centerline{
		\subfigure [\scriptsize Pitts250k-test]{
			\includegraphics[scale = 0.24,trim = 15 5 20 0,clip]{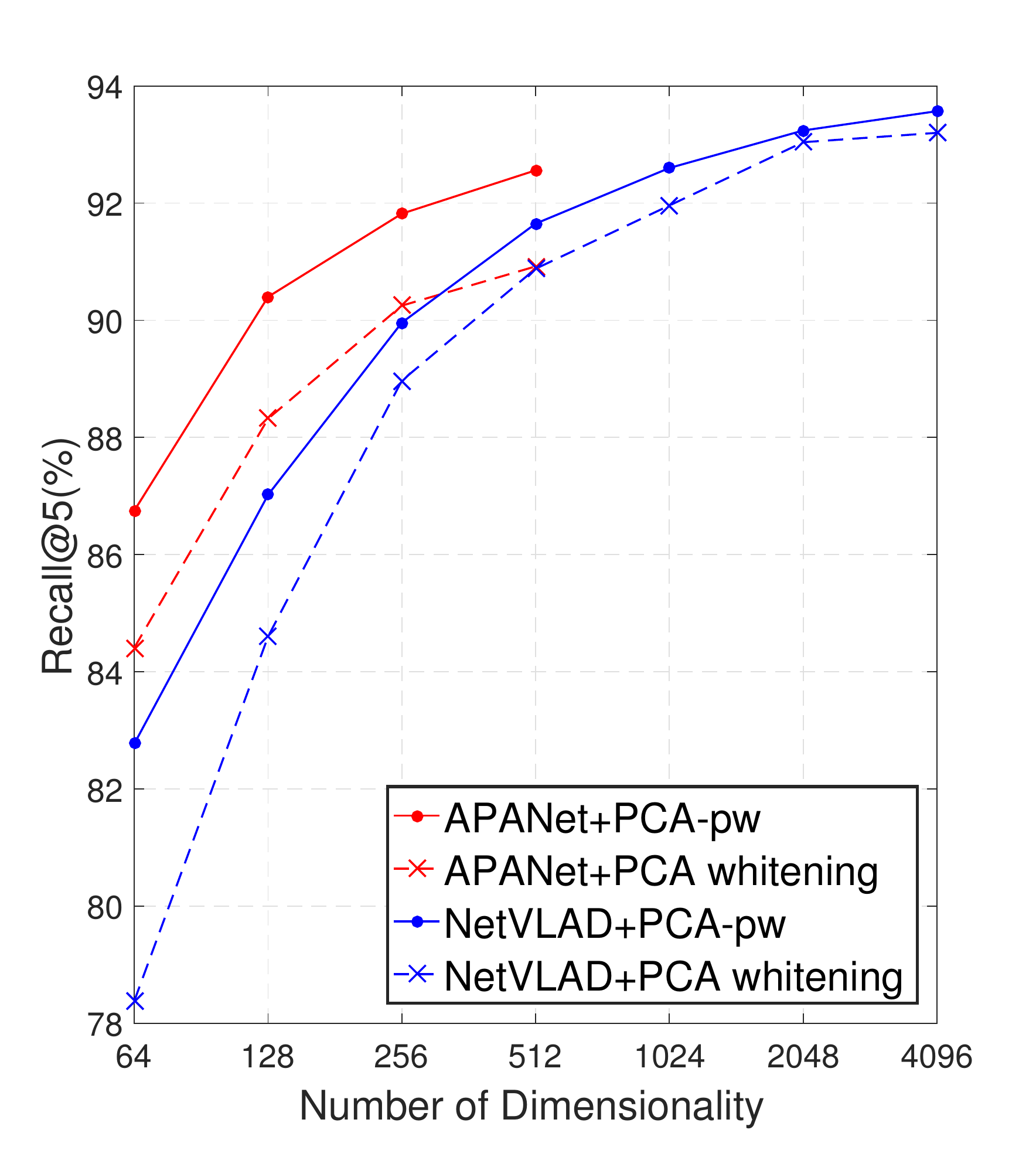}
		}
		\subfigure[\scriptsize Tokyo 24/7]{
			\includegraphics[scale = 0.24,trim = 15 5 20 0,clip]{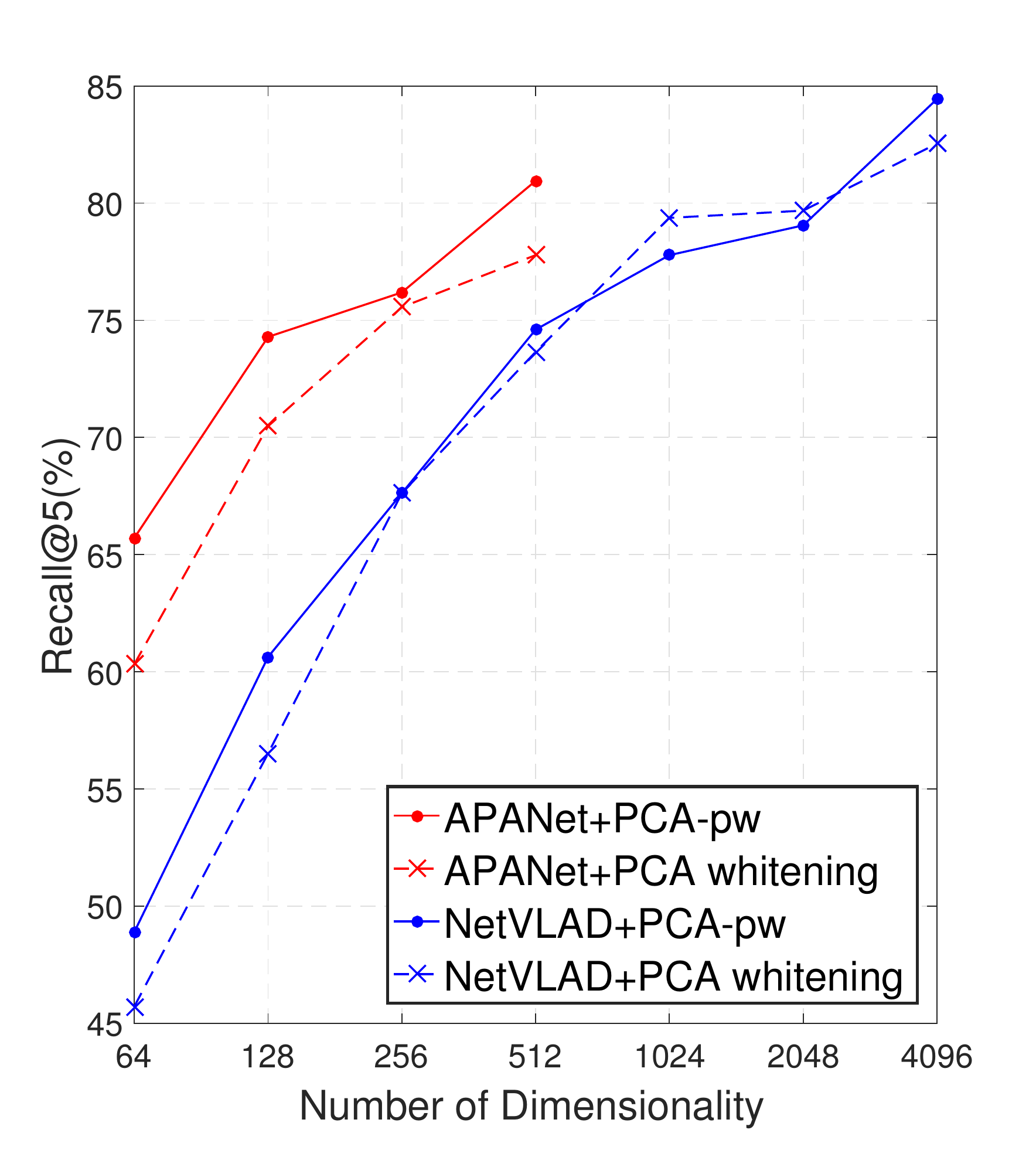}
		}
	}
	\caption{Comparison of place recognition accuracy and dimensionality.}
	\label{PTdimension}
	\end{figure}
	
    \textbf{Dimensionality reduction with PCA power whitening.}
	To further assess the performance of APANet, we present the performance of APANet and NetVLAD on varying representation dimensionalities  in Fig.~\ref{PTdimension}. We observe that   PCA-pw consistently outperforms PCA whitening for APANet representations from low to high dimensionality. Meanwhile PCA-pw improves NetVLAD on the Pitts250k-test dataset but the improvements are not so stable on the Tokyo 24/7. We speculate there may be a more suitable scaling factor $\alpha$ for NetVLAD representations on Tokyo 24/7 dataset. 
	Moreover, compared to the NetVLAD, the recall@5 curves of APANet decrease gracefully with dimensionality reduction. For similar performance, APANet representations are usually two-times more compact than NetVLAD. This phenomenon is more pronounced on the challenging Tokyo 24/7 dataset.
	
	
	
	\subsection{Instance Image Retrieval}

	To  evaluate the generalization ability of APANet, we deploy the APANet model (trained on Pitts30k-train dataset) on two standard image retrieval benchmarks, the Oxford5k \cite{philbin2007object} and Paris6k \cite{philbin2008lost} datasets. We have thorough comparisons on dimensionality with NetVLAD. 
	The results are displayed in Table~\ref{tab:image retrieval}. All the results are based on the single scale image representations and no spatial re-ranking or query expansion is adopted. We observe that when learning the whitening parameters from Pitts30k-train dataset as NetVLAD does, APANet even outperforms the 4096-D NetVLAD representations on both two datasets by 512-D representations, and it still performs well with extremely short codes.
	Further, when learning the whitening parameters from the Oxford5k or Paris6k dataset representations as conventional practices do, APANet gets consistent performance improvements from high to low dimensionality. 
	\begin{table}[h]
		\caption{Comparison with NetVLAD on image retrieval datasets. The accuracy is measured by mean average precision (mAP) and these methods are based on VGG-16 architecture. $\dagger$ denotes that the results at the column are from representations whitened on Pitts30k-train dataset and $\ddagger$ denotes Oxford5k or Paris6k.}
		\label{tab:image retrieval}
		\setlength{\abovecaptionskip}{0.cm}
		\setlength{\belowcaptionskip}{-0.cm}
		\centerline{
			\resizebox{.47\textwidth}{!}{
				\begin{tabular}{|l|c|c c|c|c c|}
					\hline
					\multirow{2}*{Dim}  & \multicolumn{3}{c|} {Oxford5k}&\multicolumn{3}{c|} {Paris6k} \\
					\cline{2-4} \cline{5-7}
					&  NetVLAD\cite{arandjelovic2016netvlad}&\multicolumn{2}{|c|} {APANet} & NetVLAD\cite{arandjelovic2016netvlad}& \multicolumn{2}{|c|} {APANet} \\
					\hline
					4096& 71.6$\dagger$ &- &-&79.7$\dagger$&-&-\\
					2048& 70.8&- &-&78.3&-&-\\
					1024& 69.2 &- &-&76.5&-&-\\
					512& 67.6 &75.1$\dagger$&\textbf{77.9$\ddagger$}  &74.9&80.2$\dagger$&\textbf{83.5$\ddagger$}\\
					256& 63.5&72.8&\textbf{75.6} &73.5&76.9&\textbf{81.7}\\
					128& 61.4 &67.3&\textbf{71.7} &69.5&74.8&\textbf{78.7}\\
					64& 51.1 &58.5&\textbf{63.9}  &63.0&70.7&\textbf{73.0}\\
					32& 42.6 &46.4&\textbf{48.7} &54.4&62.5&\textbf{63.7}\\
					16& 29.9&31.7&\textbf{33.4}  &44.9&48.3&\textbf{52.4}\\
					\hline
				\end{tabular}
			}
		}
	\end{table}
	\section{Conclusions}
	In this paper, we propose an APANet which is well-designed to overcome the challenges in place recognition task. Experiments demonstrate that APANet representations are robust to changes of viewpoint and illumination and outperform NetVLAD using the same or even lower dimensional representations. Meanwhile, APANet emerges powerful generalization ability on standard image retrieval datasets.
	In addition, the proposed PCA power whitening strategy consistently improves  performance for APANet and is applicable for other retrieval tasks as well. 
	In our future works, we will improve our APANet for instance image retrieval task.
	
	\section*{Acknowledgments}
	This work was supported by: (i) National Natural Science Foundation of China (Grant No. 61602314); (ii) Natural Science Foundation of Guangdong Province of China (Grant No. 2016A030313043); (iii) Fundamental Research Project in the Science and Technology Plan of Shenzhen (Grant No. JCYJ20160331114551175). We would also like to thank Relja Arandjelovi{\'c} and Akihiko Torii for providing data, codes, and sharing insights, and Jie Lin for insightful discussions.

\bibliographystyle{ACM-Reference-Format}
 \balance
\bibliography{sample-sigconf-MM.bbl}

\end{document}